\documentclass[letterpaper]{article} 
\usepackage{aaai2027}  
\usepackage[hyphens]{url}  
\usepackage{graphicx} 
\usepackage{natbib}  
\usepackage{caption} 
\usepackage{algorithm}
\usepackage{algorithmic}

\usepackage{amsmath,amssymb,amsfonts}
\usepackage{booktabs}
\usepackage{array}

\usepackage{newfloat}
\usepackage{listings}
\DeclareCaptionStyle{ruled}{labelfont=normalfont,labelsep=colon,strut=off} 
\floatstyle{ruled}
\newfloat{listing}{tb}{lst}{}
\floatname{listing}{Listing}

\usepackage{booktabs}

\title{RADE: Risk-Adjusted Deployment Efficiency,
\\ 
A Framework for Routing, Memory, and Escalation in Safe LLM Systems}

\author{
    Chaymaa Abbas,
    Kareem Hassani,
    Mariette Awad
}
\affiliations{
    Department of Electrical and Computer Engineering,\\
    American University of Beirut,\\
    Riad El-Solh, Beirut 1107 2020, Lebanon\\
    \{cwa07, krh10\}@mail.aub.edu, mariette.awad@aub.edu.lb
}

\begin{document}
\maketitle

\begin{abstract}
A deployed language model does not merely answer questions. It spends computation, consumes time, consults evidence, exposes memory, invites or avoids human review, and sometimes makes mistakes whose consequences are radically unequal. Yet LLM systems are still commonly compared as if the relevant object were a single model and the relevant outcome were a single benchmark score. This paper argues that the unit of analysis should instead be the deployment policy: the mechanism that decides when to retrieve, when to remember, when to call a larger model, when to verify, and when not to answer at all. We introduce risk-adjusted deployment efficiency (RADE), a system-level objective that scores the utility a deployment delivers net of inference, retry, oversight, and priced expected failure cost, while treating harms that cannot be priced as hard budgets alongside constraints on latency, privacy, and resources. The framework brings selective routing, retrieval and governed memory, efficient specialization, uncertainty estimation, abstention, and human escalation into a common account. We then examine five deployment regimes in which the same nominal LLM capability leads to different rational architectures, and propose a reporting protocol based on risk-coverage behavior, end-to-end cost, update drift, memory exposure, evidence quality, and escalation load. The central claim is modest but consequential: frontier capability should be treated as a resource to allocate, not a default state to maintain. A safe and sustainable system is not the one that always uses the strongest model; it is the one that spends capability where the reduction in risk or increase in utility justifies the cost.
\end{abstract}

\section{Introduction}

Consider a deployed assistant at nine o'clock on an ordinary morning. The first request asks it to rewrite a meeting note. The second asks for the current return policy of a supplier whose documentation changes often. The third asks whether an account action should be executed on behalf of a customer.

To a benchmark, these may all appear to be instances of language generation. To a responsible operator, they are different species of problem. The first can often be handled locally and cheaply. The second demands fresh evidence. The third may require authorization checks, a verifier, and possibly a human. The interesting engineering question is therefore not, "Which model answers all three best?" It is, "What should the system spend on each request, and what should it refuse to spend?"

That question becomes more important as LLMs move from standalone prototypes into production infrastructure. In a persistent deployment, a model is only one actor in a longer chain. An answer may depend on retrieved documents, a memory store, a tool call, a routing decision, a safety policy, a versioned model, and a person who reviews uncertain cases. A fluent completion can still be wrong because the evidence was stale, the wrong memory was retrieved, the request was routed to an underpowered model, or a model update silently changed behavior that had previously been validated. Frontier capability continues to improve, while responsible-AI evaluation and reporting remain uneven \citep{aiindex2026}. Larger and more instructable models are not uniformly more reliable, particularly when uncertainty is poorly expressed \citep{zhou2024larger}; semantic uncertainty can reveal confidently generated confabulations that surface fluency hides \citep{farquhar2024semantic}.

The usual story of progress begins with scale. Scaling laws help explain why additional parameters, data, and compute can improve average capability \citep{kaplan2020scaling,hoffmann2022chinchilla}. But deployment changes the object being optimized. A larger model may improve first-pass accuracy and simultaneously increase latency, marginal cost, memory pressure, and dependence on a remote service. A smaller model may be inexpensive per call and expensive per completed task if it triggers frequent retries and human review. A human checker may reduce severe failures and make the system unusably slow. A retrieval layer may improve freshness and introduce a new route for poisoning or stale evidence. The cheapest model call is not necessarily the cheapest answer, and the most accurate model is not necessarily the safest system.

Model-level measures such as cost per call and benchmark accuracy do not determine what a completed answer costs or how safely a deployed system behaves. This paper addresses that mismatch with a deployment-first view: we treat a deployed LLM system as a policy for allocating capability, evidence, memory, verification, and human attention. Prior work has already supplied many of the mechanisms: cascades and learned routing can shift traffic between models \citep{chen2023frugalgpt,ong2024routellm}; retrieval can externalize part of the knowledge burden \citep{lewis2020rag,borgeaud2022retro}; sparse computation and serving systems can reduce active compute and memory overhead \citep{fedus2022switch,kwon2023paged}; and uncertainty methods can help identify cases that should be rejected or escalated \citep{xiong2024uncertainty,farquhar2024semantic}. What these lines of work lack in common is not another component but a shared question: \emph{when is each component that improves an answer worth invoking, given the likelihood and consequence of that answer being wrong?}

\subsection{Statement of Contributions}
This paper makes four contributions. First, it introduces risk-adjusted deployment efficiency (RADE), a constrained net-value objective that subtracts inference, retry, human-review, and priced expected failure costs from delivered utility, while enforcing explicit budgets on harms that cannot be priced and constraints on latency, privacy, and resources. Second, it recasts routing, retrieval, memory, verification, and escalation as parts of one control policy whose task is to allocate expensive capability selectively. Third, it uses five representative deployment regimes to show why the same model can be a sensible choice in one setting and a poor choice in another. Fourth, it proposes a deployment-oriented evaluation protocol centered on risk-coverage behavior, end-to-end cost, update drift, memory governance, evidence quality, and escalation load.

\section{The Deployment Policy as the Unit of Analysis}
\label{sec:wrongunit}

The vocabulary of modern LLM evaluation is still strongly model-centric. We ask how many parameters a model has, what score it obtains, how much a token costs, or whether one model beats another on a benchmark. Those are useful facts. They are not yet a deployment decision.

Suppose a system contains two language models: a small local model and a larger general model. It also has a search index, a persistent memory store, a deterministic calculator, and a human-review queue. One could ask which language model is better, but that misses most of the policy. The real system chooses among paths. It may answer locally, retrieve first, call the larger model, verify the draft, invoke a tool, ask a clarifying question, or abstain. The object of interest is therefore not a model $m$ but a system $s$ that maps a request and its context to a sequence of actions.

This shift sounds semantic until failures become asymmetric. If the first request in our opening example is mishandled, the user may rewrite a sentence manually. If the third request is mishandled, an unauthorized action may be executed. Averaging the two into one accuracy number erases the fact that one error is cheap and the other expensive. A deployment metric must therefore know something about consequence, not merely frequency.

We represent a deployment profile as
\begin{equation}
 d = (\mathcal{D}, B, T_{\max}, P_{\max}, \boldsymbol{\epsilon}, H),
 \label{eq:profile}
\end{equation}
where $\mathcal{D}$ is the deployment distribution, $B$ is a resource budget, $T_{\max}$ is a latency constraint, $P_{\max}$ is a privacy or exposure constraint, $\boldsymbol{\epsilon}$ contains tolerated probabilities for critical failure classes, and $H$ describes the available human oversight. The important feature of $d$ is that it describes the world around the model. The same model placed in a different world can rationally lead to a different system.
\section{Risk-Adjusted Deployment Efficiency}
\label{sec:rade}

A useful accounting identity follows from a simple observation: a deployment pays a price for more than inference. It pays when a weak answer triggers a retry. It pays when a person must review the result. It pays, sometimes overwhelmingly, when a consequential failure occurs.

For each failure class $k$, a factual error, an unauthorized action, a privacy disclosure, an unsafe clinical suggestion, or another domain-specific event, let $p_k(s,d)$ be its probability per request under system $s$ and deployment profile $d$. Failure classes differ in whether their consequences can be priced. We therefore partition them into $\mathcal{K}_{\$}$, classes whose consequence $L_k(d)$ admits a defensible common unit (for example, a refund issued in error, wasted staff time, or a contractual penalty), and $\mathcal{K}_{\mathrm{sev}}$, classes that admit only a severity ordering (for example, a severe clinical error or the disclosure of sensitive personal data). For $k \in \mathcal{K}_{\$}$, $L_k(d)$ measures the damage caused by the failure itself, excluding the value of the task that was not delivered, which is accounted for separately below. We define the expected risk-adjusted cost per request as
\begin{equation}
 C_{\mathrm{RA}}(s,d)
 = C_{\mathrm{infer}}
 + C_{\mathrm{retry}}
 + C_{\mathrm{review}}
 + \sum_{k \in \mathcal{K}_{\$}} p_k(s,d)L_k(d).
 \label{eq:cra}
\end{equation}

The distinction in Equation~\ref{eq:cra} is elementary but often lost in model comparisons. Inference cost is paid directly; failure cost is paid probabilistically. In a low-consequence task, the former may dominate. In a high-consequence task, a small change in the latter can dwarf everything else. Classes in $\mathcal{K}_{\mathrm{sev}}$ are deliberately absent from the sum: adding a severity score to a monetary cost would produce a number with no meaningful unit.

Let $U(s,d)$ denote the expected utility per request delivered by requests that end in a correct completion under the deployment profile. A request counts toward $U$ whether it is completed on the first path or after escalation to a stronger model or a human reviewer; the cost of that escalation is already carried by $C_{\mathrm{retry}}$ or $C_{\mathrm{review}}$. Incorrect outputs and requests that end without a correct completion contribute nothing to $U$; the damage a failure causes enters only through $L_k$. A failure therefore registers in two places for two distinct reasons, once as value not delivered and once as damage done, but no single loss is counted twice. We define risk-adjusted deployment efficiency as a net value,
\begin{equation}
 \mathrm{RADE}(s,d) = U(s,d) - \lambda\, C_{\mathrm{RA}}(s,d),
 \label{eq:rade}
\end{equation}
where $\lambda > 0$ converts cost into utility units ($\lambda = 1$ when utility is itself monetary). The deployment problem is to maximize Equation~\ref{eq:rade} subject to hard conditions such as
\begin{align}
 p_k(s,d) &\leq \epsilon_k && \forall k \in \mathcal{K}_{\mathrm{critical}}, \label{eq:con-fail}\\
 \mathbb{E}[T(s,d)] &\leq T_{\max},\\
 P(s,d) &\leq P_{\max},\\
 C_{\mathrm{resource}}(s,d) &\leq B, \label{eq:con-res}
\end{align}
where $T(s,d)$ is the end-to-end latency of a request, $P(s,d)$ is the extent to which private data is sent or retained beyond the boundary the deployment permits, for example by routing it to an external service or storing it in persistent memory, and $C_{\mathrm{resource}}(s,d)$ is the computational capacity the system requires, such as accelerator compute and memory; each is bounded by the corresponding element of the deployment profile $d$. The privacy constraint is therefore a structural limit on exposure, distinct from the probability of a disclosure failure, which Equation~\ref{eq:con-fail} bounds. The set $\mathcal{K}_{\mathrm{critical}}$ contains every class in $\mathcal{K}_{\mathrm{sev}}$ together with any priced class whose probability must also be capped. The damage caused by classes in $\mathcal{K}_{\mathrm{sev}}$ therefore never enters the objective. A severe harm is treated as a budget to respect, not a cost to trade. We keep the word efficiency in its allocative sense: a system that maximizes Equation~\ref{eq:rade} inside the constraints cannot reallocate capability, evidence, or oversight to deliver more net value.
The difference form is chosen over a ratio such as $U / C_{\mathrm{RA}}$ for two reasons. First, the zero point and scale of utility are conventions. Replacing $U$ with $aU + b$ for $a > 0$ leaves the ranking produced by Equation~\ref{eq:rade} unchanged once $\lambda$ is rescaled by $a$, whereas a shift $b$ in the zero point can reorder systems ranked by a ratio. Second, a ratio rewards systems that do very little very cheaply. Consider system $s_1$, which completes 20\% of requests correctly at \$0.01 per request, and system $s_2$, which completes 90\% at \$0.10, with each correct completion worth \$1 and no priced failures. The ratio scores $s_1$ at 20 and $s_2$ at 9; the net value scores $s_1$ at 0.19 and $s_2$ at 0.80. Only the second ranking matches the choice an operator would make. When no exchange rate is agreed, $\lambda$ should be treated as a reporting parameter: results should be shown across a range of values, or as the frontier of $U$ against $C_{\mathrm{RA}}$ among feasible systems, so that readers can see whether a ranking depends on the choice.

What a completed answer costs is captured by a secondary measure. Let $A(s,d)$ denote the probability that a request ends in a correct completion. The price of an answer is
\begin{equation}
 \pi(s,d) = \frac{C_{\mathrm{RA}}(s,d)}{A(s,d)},
 \label{eq:price}
\end{equation}
the expected risk-adjusted spend per correct completion. Because $\pi$ is itself a ratio, it inherits the pathology above: in the example, $s_1$ pays \$0.05 per correct answer and $s_2$ pays about \$0.11. It is therefore descriptive rather than an objective, and should be reported only alongside $A(s,d)$ or compared among systems that meet a declared minimum completion rate.

RADE is not proposed as a new training loss. It is an accounting frame for deployment. The constraints in Equations~\ref{eq:con-fail}-\ref{eq:con-res} come first: a high net value should not compensate for violating a hard privacy or severe-failure bound. Only among systems inside the declared safety envelope does the comparison in Equation~\ref{eq:rade} become meaningful.

This view explains several choices that appear contradictory when judged at the model level. A cheap model that frequently induces retries can cost more than a stronger model on hard queries. A frontier model used for every request can be wasteful if most requests are routine. Human review can be valuable precisely because it is expensive: it should be reserved for cases in which the expected reduction in harm justifies the delay and labor. Retrieval can save the cost of retraining for new facts, but it must earn that saving by providing evidence that is fresh enough and trustworthy enough to improve the decision.

Work on model cascades already points in this direction. FrugalGPT shows that heterogeneous model combinations can improve the cost-quality frontier \citep{chen2023frugalgpt}; RouteLLM learns to route between stronger and weaker models using preference data \citep{ong2024routellm}. RADE adds a condition that becomes crucial in deployment: the router should care not only about the probability of being wrong, but also about the consequence of being wrong.
\section{Deployment Regimes and Binding Constraints}
\label{sec:regimes}

The easiest way to see why a universal "best model" is a poor abstraction is to walk the same technology through different regimes. Table~\ref{tab:regimes} summarizes five representative regimes. The point is not that all deployments fit neatly into these categories. The point is that the binding constraint changes.

\begin{table*}[t]
\centering
\small
\caption{Five representative deployment regimes. The relevant "failure budget" concerns the tolerated probability and severity of consequential failure, not ordinary benchmark error.}
\label{tab:regimes}
\begin{tabular}{@{}p{2.5cm}p{2.4cm}p{2.5cm}p{2.6cm}p{4.4cm}@{}}
\toprule
\textbf{Regime} & \textbf{Infrastructure} & \textbf{Binding constraint} & \textbf{Failure budget} & \textbf{Rational deployment priority} \\
\midrule
Research assistant & Cloud, multi-model & Marginal cost and evidence freshness & Moderate; users can often verify & Route routine work cheaply; spend on retrieval and verification for factual claims. \\
Enterprise on-premise & Local GPUs / private cloud & Privacy, hardware capacity, maintainability & Low for disclosure and policy violations & Prefer bounded local models, private retrieval, governed memory, and controlled escalation. \\
High-volume action support & Cloud, tool-using & Unauthorized action and review load & Very low for irreversible actions & Separate language generation from authorization; verify before execution. \\
Offline clinical assistant & Edge / hospital network & Privacy, latency, evidence staleness & Extremely low for severe clinical errors & Constrain scope, surface evidence freshness, abstain aggressively, escalate to a clinician. \\
General knowledge work & Cloud plus optional local & Heterogeneous tasks and long-horizon drift & Task-dependent & Route dynamically, govern persistent memory, and regression-test behavior across updates. \\
\bottomrule
\end{tabular}
\end{table*}

In the research-assistant regime, abstention is often cheap. A user can ask a follow-up, open the cited source, or rephrase the request. The system can therefore trade some automation for lower cost. In the enterprise regime, privacy may dominate: sending every query to a remote frontier model can be unacceptable even if that model is more capable on average. In high-volume support, the decisive boundary is not between a good sentence and a bad sentence; it is between text generation and authorization to act. An assistant may be permitted to draft a refund explanation while being forbidden to issue the refund without a separate check.

The clinical regime makes the point most sharply. Local execution can improve privacy and availability, but a perfectly private model with stale evidence can still be dangerous. The binding risk may therefore shift from disclosure to freshness. In general knowledge work, no single risk dominates for long: the system must cope with varied tasks, persistent memory, changing models, and repeated updates.

A deployment-first framework asks, in each regime, what the system must know before it spends capability. Model size becomes one control knob among several.

\section{Selective Computation as a Safety Policy}
\label{sec:control}

Selective computation is usually introduced as an efficiency technique: send easy requests down a cheap path and difficult requests down an expensive one. That description is incomplete. A production router is also a safety policy because it decides which requests are allowed to receive which level of competence, evidence, and oversight.

\begin{figure*}[t]
\centering
\includegraphics[width=0.92\textwidth]{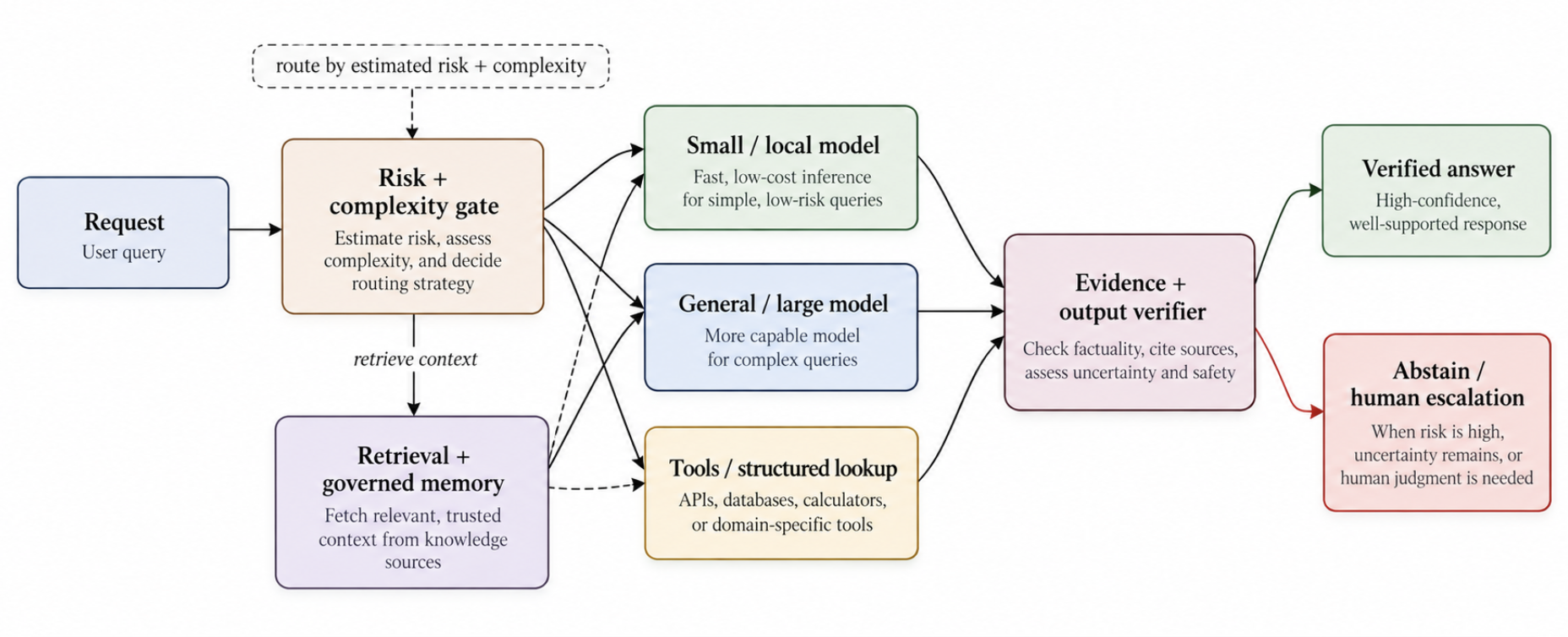}
\caption{A deployment-first control architecture. A low-cost gate estimates complexity and consequence; retrieval and governed memory provide context; heterogeneous execution paths produce candidate outputs; and verification determines whether a verified answer is released or the case is withheld and escalated. The figure is architectural rather than an empirical benchmark result.}
\label{fig:stack}
\end{figure*}

Figure~\ref{fig:stack} shows the architecture implied by this view. A low-cost gate first estimates something richer than difficulty. It asks whether the request is routine or high consequence, whether current evidence is required, whether private memory is relevant, whether a tool call would create an irreversible effect, and whether the system can recognize the limits of its own competence. From there the request is routed to a small model, a larger model, or a tool, drawing on retrieval and memory where needed. The final decision is not simply which model generates the answer, but whether that answer is backed by enough evidence and verification to be released at all. 

\subsection{Severity-Aware Routing}

Learned routing and cascades already show that not every query requires the same model \citep{chen2023frugalgpt,ong2024routellm}. A safety-aware router adds two requirements. First, it should route partly on failure severity, not only predicted answer quality. Second, escalation should remain possible after generation. A cheap path may look adequate at the start and become inadequate when retrieval fails, evidence conflicts, or a verifier detects uncertainty.

Several efficiency techniques can be understood through the same lens. Sparse expert models separate total capacity from active computation \citep{fedus2022switch}. State-space architectures such as Mamba explore alternatives to quadratic attention for long sequences \citep{gu2023mamba}. Speculative decoding drafts with a cheaper model and verifies with a stronger one \citep{leviathan2023speculative}. The internal mechanisms differ, but the systems principle is common: expensive computation should be conditional rather than automatic.

\subsection{Retrieval and Evidence}

The second request in our opening example asks for a policy that changes over time. Here, parametric memory is a poor place to put operational truth. Retrieval-augmented generation and retrieval-augmented pretraining show how external corpora can supply knowledge without encoding every update into model weights \citep{lewis2020rag,borgeaud2022retro}. The deployment advantage is not merely freshness. Retrieval creates an evidence path that can be inspected, versioned, and, at least in principle, verified.

But retrieval is not proof. A stale document, a poisoned index, or a badly matched passage can produce an answer that looks more grounded because it contains citations. Retrieved evidence should therefore inform verification as well as generation: evidence is an input to judgment, not a substitute for it.

\subsection{Infrastructure and Safety Margins}

Systems work is often treated as an implementation detail that begins after model evaluation. In deployment, it can change which safety measures remain available. Long-context serving consumes substantial key-value-cache memory; PagedAttention improves memory management and serving throughput by reducing fragmentation and enabling more flexible allocation \citep{kwon2023paged}. This matters because a system running close to its resource limit may respond to load by truncating context, disabling a verifier, shrinking retrieval, or creating queues that defeat latency guarantees. Resource headroom is therefore not merely an efficiency concern. It is part of the margin within which safety mechanisms continue to function.

\section{Memory, Continual Learning, and Governance}
\label{sec:memory}

A deployed assistant is expected to learn in a peculiar sense of the word. It should remember a user's preference tomorrow, reflect a changed policy next week, and still retain capabilities that were validated last month. These demands are often grouped under "continual learning," but they happen on different timescales and should not all be solved by changing model weights.

The danger of collapsing these timescales is familiar. Continual fine-tuning can produce catastrophic forgetting or distribution-specific capability loss \citep{french1999catastrophic,luo2023catastrophic,kotha2024forgetting}. Even successive releases of general-purpose models can exhibit behavioral drift despite improvements in aggregate capability \citep{chen2023chatgpt}. A system that updates frequently without remembering what it used to do is not learning in the operational sense; it is changing.

Complementary Learning Systems theory offers a useful analogy: fast episodic adaptation and slow semantic consolidation should occur on different timescales \citep{mcclelland1995cls,kumaran2016cls}. For LLM deployment, the analogy suggests three layers.

\paragraph{Fast memory.}
Recent facts, user preferences, task state, and organization-specific knowledge can live outside the base model and be retrieved only when relevant. Systems such as MemGPT make this separation explicit \citep{packer2023memgpt}. The advantage is reversibility: a memory item can be corrected or deleted without retraining the model.

\paragraph{Slow consolidation.}
Stable patterns may eventually justify a parameter update, but such updates should be less frequent and surrounded by regression tests. Classical continual-learning methods such as Elastic Weight Consolidation and Synaptic Intelligence illustrate the general principle of protecting important representations while incorporating new information \citep{kirkpatrick2017ewc,zenke2017synaptic}. Whether these exact methods are the right mechanism for modern LLMs is less important here than the architectural lesson: fast local change need not imply immediate global change.

\paragraph{A governance layer between the two.}
Memory is useful because it preserves information across time; that is also why it can become dangerous. Training-data extraction has shown that memorized text can be recoverable from language models \citep{carlini2021extracting}, and recent work demonstrates black-box extraction attacks against agent memory \citep{wang2025privacy}. Persistent memory should therefore be treated as a governed data store. It needs write policies, retrieval scopes, provenance, retention rules, expiry, deletion, and adversarial tests. A memory subsystem with no policy is not personalization; it is uncontrolled persistence.

This reframing changes the practical meaning of continual learning. When new information arrives, the safest first move may be to update retrieval or episodic memory, observe how the information is used, and consolidate only what proves stable. The system gains plasticity without paying the full stability cost on every update.

\section{Uncertainty, Abstention, and Escalation}
\label{sec:abstention}

A capable model can produce an answer to almost any well-formed prompt. A reliable system must sometimes resist that capability.

For high-consequence deployment, average correctness is not enough. The system must distinguish cases in which it is competent from cases in which it should seek more evidence or hand control elsewhere. Larger models can remain overconfident or become less reliable along dimensions that aggregate benchmarks do not reveal \citep{zhou2024larger}. RLHF can also amplify verbalized overconfidence through reward-model biases \citep{leng2025rlhfcalibration}. This is why uncertainty estimation matters most at the boundary between action and abstention.

Let $u(x)$ denote an uncertainty or risk score for request $x$. Suppose the system answers only when $u(x)\leq\tau$. We can then define coverage and selective risk as
\begin{align}
 \kappa(\tau) &= \Pr(u(x)\leq\tau),\\
 R_{\mathrm{sel}}(\tau) &= \mathbb{E}[\ell(\hat{y},y)\mid u(x)\leq\tau].
 \label{eq:selective}
\end{align}
A useful score should reduce selective risk as coverage falls, especially on shifted or high-severity subsets. Free-form generation makes this difficult because many strings can express the same underlying claim. Semantic entropy addresses part of the problem by measuring uncertainty over meaning clusters and improves rejection behavior for confabulation-prone questions \citep{farquhar2024semantic}. Other work on confidence elicitation shows that verbal or probability-like confidence is not automatically calibrated \citep{xiong2024uncertainty}.

Abstention is not synonymous with refusal. The system has a repertoire of non-answers: ask for missing information, retrieve another source, call a more capable model, invoke a deterministic tool, request human review, or block an irreversible action. The operational problem is therefore not "How often should the model refuse?" but "Which escalation should follow which uncertainty?"

The economics return here. If a first-pass answer costs $c_i$, triggers a retry with probability $q_r$ at cost $c_r$, triggers human review with probability $q_h$ at cost $c_h$, and incurs failure class $k$ with probability $p_k$, then the expected risk-adjusted cost per request, the request-level form of Equation~\ref{eq:cra}, is
\begin{equation}
 c_i + q_r c_r + q_h c_h + \sum_{k \in \mathcal{K}_{\$}} p_k L_k.
 \label{eq:completedcost}
\end{equation}
A more expensive first-pass model can be cheaper overall if it sufficiently reduces retries, review, and priced failures. It can also be the only admissible choice if it is the one that keeps the classes in $\mathcal{K}_{\mathrm{sev}}$ within their budgets in Equation~\ref{eq:con-fail}. A cheap model plus a good router and verifier can also dominate an expensive model used blindly. What matters is the policy, not the sticker price.

\section{Bounded Specialization}
\label{sec:specialize}

Efficiency arguments often end with a recommendation to use smaller models. That conclusion is too easy. Smaller models are attractive in privacy-sensitive, on-premise, and edge settings because they reduce serving requirements. Compute-optimal training already shows that parameter count alone is an incomplete measure of capability \citep{hoffmann2022chinchilla}; domain-adaptive pretraining shows that in-domain data can produce large gains without requiring a universally larger model \citep{gururangan2020dont}.

But specialization buys competence by narrowing the world in which competence is expected. A domain model may be excellent inside its distribution and brittle just outside it. The safer pattern is therefore \emph{bounded specialization}: define the region in which the smaller model is trusted, monitor for shift, and preserve an escalation path to a broader model or a person.

The same logic applies to systems optimization. Quantization, pruning, batching, and cache management should not be evaluated only by throughput. An optimization that produces a tiny average accuracy loss can still be harmful if that loss is concentrated in rare high-severity cases. A system can become faster and less efficient in the RADE sense, or faster and inadmissible if the loss pushes a class in $\mathcal{K}_{\mathrm{sev}}$ past its budget in Equation~\ref{eq:con-fail}. Safety evaluation must therefore be stratified by consequence, not only by frequency.

\section{Evaluation and Reporting Protocol}
\label{sec:evaluation}

Once the unit of analysis moves from model to policy, the evaluation report must change with it. A deployment paper should let a reader reconstruct the operational trade-off: what was automated, what was escalated, what the system spent, what changed after updates, and where the dangerous failures occurred.

Table~\ref{tab:metrics} gives a minimum reporting set. None of the rows is exotic. Their importance comes from reporting them together.

\begin{table*}[t]
\centering
\small
\caption{A minimum reporting set for risk-adjusted evaluation of deployed LLM systems.}
\label{tab:metrics}
\begin{tabular}{@{}p{3.0cm}p{5.1cm}p{6.5cm}@{}}
\toprule
\textbf{Dimension} & \textbf{Report} & \textbf{Question it answers} \\
\midrule
Task utility & Accuracy / task success / application utility on a representative deployment distribution & Does the system perform useful work, including on important subgroups or scenarios? \\
Selective reliability & Risk-coverage curve, abstention rate, uncertainty or calibration quality & Does the system know which cases to escalate? \\
Failure severity & Error taxonomy and frequency by severity class & Are rare consequential failures hidden by many harmless successes? \\
Resource cost & Tokens, active-compute proxy, memory footprint, latency, and monetary cost where available & What does a completed task actually consume? \\
Oversight load & Human-review rate, retry rate, escalation destination, time-to-resolution & How much hidden labor is required to make the system reliable? \\
Update stability & Regression-suite delta, forgetting rate, policy drift after updates & Does improvement in one version silently break validated behavior elsewhere? \\
Memory governance & Write/read policy, retention, deletion test, adversarial extraction or leakage rate & What persistent information can be stored, retrieved, and exposed? \\
Evidence quality & Retrieval recall/precision, source freshness, verifier pass/fail behavior & Are answers supported by evidence that is relevant and current? \\
\bottomrule
\end{tabular}
\end{table*}

\subsection{End-to-End Evaluation}

Component metrics can be individually strong while the system fails. A router may predict average model preference accurately and still send exactly the wrong high-severity cases to the cheap path. A verifier may have a strong aggregate discrimination score and still miss the small class of errors that triggers irreversible actions. End-to-end evaluation must therefore sample from the deployment distribution and include adversarial, shifted, and high-severity slices.

\subsection{Risk-Coverage and Cost-Risk Curves}

Any selective system has a threshold somewhere, even if the threshold is hidden in a prompt or business rule. Reporting risk as a function of coverage reveals whether the system can give up automation gracefully as the required safety level rises. Cost-risk curves play the same role for routing and verification: they show whether an additional model call, retrieval step, or human check buys enough risk reduction to justify its cost.

\subsection{Update Stability and Behavioral Drift}

Static evaluation is inadequate when models, indexes, tools, and policies change repeatedly. Each update should be tested against a persistent regression set and a fresh distribution-shift set. For discrete, safety-relevant outcomes, a simple behavioral drift statistic between consecutive versions $t$ and $t+1$ of the system is
\begin{equation}
 D_{t,t+1}=\Pr_{x\sim\mathcal{D}}\left[g_t(x)\neq g_{t+1}(x)\right],
 \label{eq:drift}
\end{equation}
where $g_t$ maps a request to the outcome chosen by version $t$, such as answer, abstain, escalate, or authorize action, and the probability is estimated in practice on the regression set. A nonzero value is not itself a failure, since an update may change outcomes for the better; it identifies the requests that need inspection. The statistic also cannot detect a change in the content of an answer when the outcome type stays the same, so it should be paired with task-specific scoring of correctness. Its purpose is narrower: to make silent policy change visible.

\section{Design Trade-offs}
\label{sec:tradeoffs}

A framework is useful when it clarifies a trade-off, not when it pretends to abolish one.

\paragraph{Stability and plasticity.}
Fast adaptation is valuable, yet shared-parameter updates can damage unrelated behavior \citep{french1999catastrophic,luo2023catastrophic}. Separating fast episodic updates from slower consolidation reduces the tension but does not remove it.

\paragraph{Memory and privacy.}
Persistent memory improves continuity because it preserves information that would otherwise be forgotten. The same persistence creates the exposure risk. Memory extraction results therefore make access policy part of the threat model rather than a user-interface detail \citep{wang2025privacy}.

\paragraph{Capability and efficiency.}
A smaller or quantized model reduces resource cost and usually narrows the region in which it should be trusted. The right question is not whether the small model can replace the large one everywhere, but whether the system can recognize the region in which the cheaper path remains inside the declared failure budget.

\paragraph{Automation and oversight.}
Human review can reduce some failure probabilities while adding delay, labor, and new opportunities for error. A system that sends most requests to people may be safe and economically pointless. The objective is not maximum automation; it is oversight that either earns its cost in net value or is needed to satisfy the constraints, and no more.

\paragraph{Freshness and attack surface.}
Retrieval and tools provide current information and useful actions by extending the system beyond the model. Every extension also enlarges the dependency graph. Source integrity, retrieval poisoning, tool authorization, and fallback behavior therefore belong in the same evaluation report as answer quality.

\section{Implications for Research and Standards}
\label{sec:implications}

The deployment-first view changes what counts as an informative result.

For architecture research, parameter count and benchmark score should be joined by active computation, memory footprint, and the risk profile at the chosen operating point. Selective computation becomes especially valuable when it is conditioned not merely on difficulty but on consequence and uncertainty.

For continual-learning research, improvement on the newest distribution should be reported together with behavior retained from earlier validated versions. Memory systems should state their write, retrieval, retention, and deletion semantics and be tested against extraction attacks.

For uncertainty research, correlation between confidence and correctness is only an intermediate result. An uncertainty score becomes operationally meaningful when it sends the right cases to the right escalation path at an acceptable coverage and oversight cost. Risk-coverage curves should therefore be interpreted alongside the consequence model of the target deployment.

For sustainability research, reliability is part of the resource story. A cheap model that causes frequent retries, review, or incidents may consume more resources than a stronger model used selectively. Conversely, a routing-and-verification policy can make frontier capability available without paying for it on every request.

For standards and certification, the framework suggests questions that are concrete enough to audit: What are the declared failure classes and budgets? Which requests may trigger autonomous action? How fresh must evidence be? Which component owns the decision to abstain? What memory can be written and deleted? What happens when an update changes the answer/escalate policy? These questions are more operationally meaningful than the undifferentiated claim that a model is "safe".

\section{Ethical Considerations}
\label{sec:ethics}

RADE is a ledger, and every ledger is kept from someone's point of view. The formalism in Section~\ref{sec:rade} does not by itself say whose costs are counted, whose benefits are maximized, or who decides where its lines fall. Left unstated, the default is the operator's point of view, and that default is where most of the ethical risk lies.

\paragraph{Whose harm counts.}
If $L_k$ records only what a failure costs the operator, such as refunds, fines, or legal exposure, then harm borne by users and third parties disappears from Equation~\ref{eq:cra}. A deployment can then look efficient precisely because the people it harms are cheap to compensate or have no claim at all. We therefore read $L_k$ as the damage to everyone a failure affects, not as operator liability, and $U$ as the value delivered to the people the system serves, for which revenue is at best a proxy. The examples of priced classes in Section~\ref{sec:rade} (a refund issued in error, wasted staff time, a contractual penalty) are operator-side costs chosen because they are easy to price; under this reading they are parts of $L_k$, not the whole of it, and $L_k$ must also include what the failure costs the affected user or third party, such as time lost correcting it or a loss incurred by acting on it. Payments that only move money between parties, such as an erroneous refund or a penalty paid to a client, should count at the net damage they reflect rather than at face value, since counting them at face value would record a transfer as a loss. We also read the requirement in Section~\ref{sec:rade} that a priced class admit a "defensible" common unit strictly: a unit is not defensible merely because a number can be attached to the harm. Where damage falls on people who did not accept the risk, or can be priced only by assumption, the class belongs in $\mathcal{K}_{\mathrm{sev}}$ by default. The partition then carries ethical weight: harms that should not be bought are governed by budgets that act as side constraints, not as prices.

\paragraph{Who sets the parameters.}
The partition between $\mathcal{K}_{\$}$ and $\mathcal{K}_{\mathrm{sev}}$, the losses $L_k$, the budgets $\epsilon_k$, and the exchange rate $\lambda$ determine most of what RADE will recommend, and the operator has an interest in each: looser budgets and lower losses both make more deployments look acceptable. These parameters should therefore not be set by the operator alone. Depending on the domain, they can come from sector regulation, as in clinical settings, from external oversight by independent third parties \citep{raji2022outsider}, or from the participation of the communities that bear the risk \citep{birhane2022power}. At a minimum, they should be published with their justification, so that the choices behind a favorable score can be contested.

\paragraph{Averages hide who bears the risk.}
The probabilities $p_k$ are averages over the deployment distribution $\mathcal{D}$. A system can satisfy every budget on average while a particular group experiences a much higher failure rate; language models are already known to perform worse for some languages and social groups \citep{weidinger2022taxonomy}. The budgets in Equation~\ref{eq:con-fail} should therefore hold when $\mathcal{D}$ is restricted to each affected group, not only over $\mathcal{D}$ as a whole. A related risk is specific to value-maximizing deployment. When utility is measured in money, it tends to track willingness to pay, and a policy chosen to maximize Equation~\ref{eq:rade} will tend to spend more capability on requests worth more. Differences in speed or convenience may be a legitimate commercial choice; differences in safety are not. The budgets for $\mathcal{K}_{\mathrm{sev}}$ must hold for low-value and high-value requests alike.

\paragraph{Costs that move rather than disappear.}
Some savings in the ledger are transfers. Abstention costs the operator little, but the unfinished work moves to the user. That is acceptable where users can verify and follow up, as in the research-assistant regime of Section~\ref{sec:regimes}, and less so where they cannot. Escalation moves work onto human reviewers, whose attention is finite and who tend to over-rely on automated suggestions, an effect observed in experts as well as novices \citep{parasuraman2010complacency}. The oversight-load row of Table~\ref{tab:metrics} exists to keep this labor visible.

\paragraph{A score is not a license.}
A favorable RADE does not show that a harm is acceptable; it shows that the assumptions about harm have been written down. The framework's ethical value lies in that disclosure. An explicit ledger can be audited, challenged, and corrected, while an implicit one cannot. Treating the score as a justification rather than a disclosure would fall into a version of what \citet{selbst2019fairness} call the formalism trap: treating a question that remains contested in its social context as if a formal quantity could settle it.

\section{Scope and Limitations}
\label{sec:limitations}

This paper is a conceptual and integrative contribution rather than a new empirical benchmark. RADE does not solve the moral or methodological problem of assigning costs to harm. It makes the problem explicit instead: harms without a defensible common unit are placed in $\mathcal{K}_{\mathrm{sev}}$ and governed by probability budgets rather than being forced into Equation~\ref{eq:cra}. The partition itself, the budgets $\epsilon_k$, and the exchange rate $\lambda$ remain value judgments. Some consequences are non-monetary, delayed, difficult to compare, or distributed across people, and these choices should be declared and justified rather than treated as technical defaults.

The five deployment regimes are illustrative, not exhaustive. They do not cover every domain, organizational form, or cultural context. The framework also relies on routers, uncertainty estimates, verifiers, and risk labels that can fail under distribution shift. No single gate should therefore be interpreted as proof of safety. High-severity applications still require independent controls and human accountability.

Finally, resource and latency comparisons depend on hardware and serving assumptions. Empirical applications of RADE should report enough infrastructure detail for the cost terms to be interpreted correctly. The framework is intended to make those assumptions visible, not to hide them behind one universal score.

\section{Conclusion}

Return to the three requests from the beginning. The meeting note, the changing supplier policy, and the customer action do not deserve the same path merely because all three can be written as prompts. One needs inexpensive language generation. One needs evidence. One needs control.

That is the broader point. A deployed LLM system is not a model waiting to answer; it is a policy for deciding how much computation, evidence, memory, verification, and human attention a request deserves. RADE formalizes the choice of that policy as a risk-adjusted allocation problem. It asks the system to satisfy its declared safety constraints first and then to maximize the net value of useful work after operational and priced expected failure costs.

Under this view, scale remains valuable, but its role changes. Frontier capability becomes something to invoke deliberately when its expected reduction in risk or increase in utility justifies the marginal cost. Smaller models, retrieval, memory, tools, verifiers, and human review become complementary instruments rather than competitors in a model leaderboard. The measure of progress is no longer capability per parameter, nor even capability per token. It is the quality of the decisions a system can make about when to spend capability, when to seek evidence, and when to stop.

\bibliography{aaai2027}

\end{document}